\documentclass[10pt,twocolumn,letterpaper]{article}

\usepackage{iccv}
\usepackage{times}
\usepackage{epsfig}
\usepackage{graphicx}
\usepackage{mwe}
\usepackage{amsmath}
\usepackage{amssymb}
\usepackage{algorithm}
\usepackage{algpseudocode}
\usepackage{dblfloatfix} 
\usepackage{float}
\usepackage{subcaption}

\usepackage{rotating}

\usepackage[pagebackref=true,breaklinks=true,letterpaper=true,colorlinks,bookmarks=false]{hyperref}
\usepackage{cleveref}

\newcommand{\mycomment}[1]{}
\iccvfinalcopy 

\def\iccvPaperID{42} 

\ificcvfinal\fi

\begin{document}

\title{Multimodal Contrastive Learning and Tabular Attention for Automated Alzheimer's Disease Prediction}

\author{Weichen Huang\\
St Andrew's College\\
Dublin, Ireland\\
{\tt\small w.huang@students.st-andrews.ie}
}

\maketitle
\ificcvfinal\thispagestyle{empty}\fi

\begin{abstract}
  Alongside neuroimaging such as MRI scans and PET, Alzheimer's disease (AD) datasets contain valuable tabular data including AD biomarkers and clinical assessments. Existing computer vision approaches struggle to utilize this additional information. To address these needs, we propose a generalizable framework for multimodal contrastive learning of image data and tabular data, a novel tabular attention module for amplifying and ranking salient features in tables, and the application of these techniques onto Alzheimer's disease prediction. Experimental evaulations demonstrate the strength of our framework by detecting Alzheimer's disease (AD) from over 882 MR image slices from the ADNI database. We take advantage of the high interpretability of tabular data and our novel tabular attention approach and through attribution of the attention scores for each row of the table, we note and rank the most predominant features. Results show that the model is capable of an accuracy of over 83.8\%, almost a 10\% increase from previous state of the art. 
\end{abstract}

\section{Introduction}

\begin{figure*}[t!]
  \begin{minipage}[t]{.45\textwidth}
      \centering
      \includegraphics[width=\linewidth]{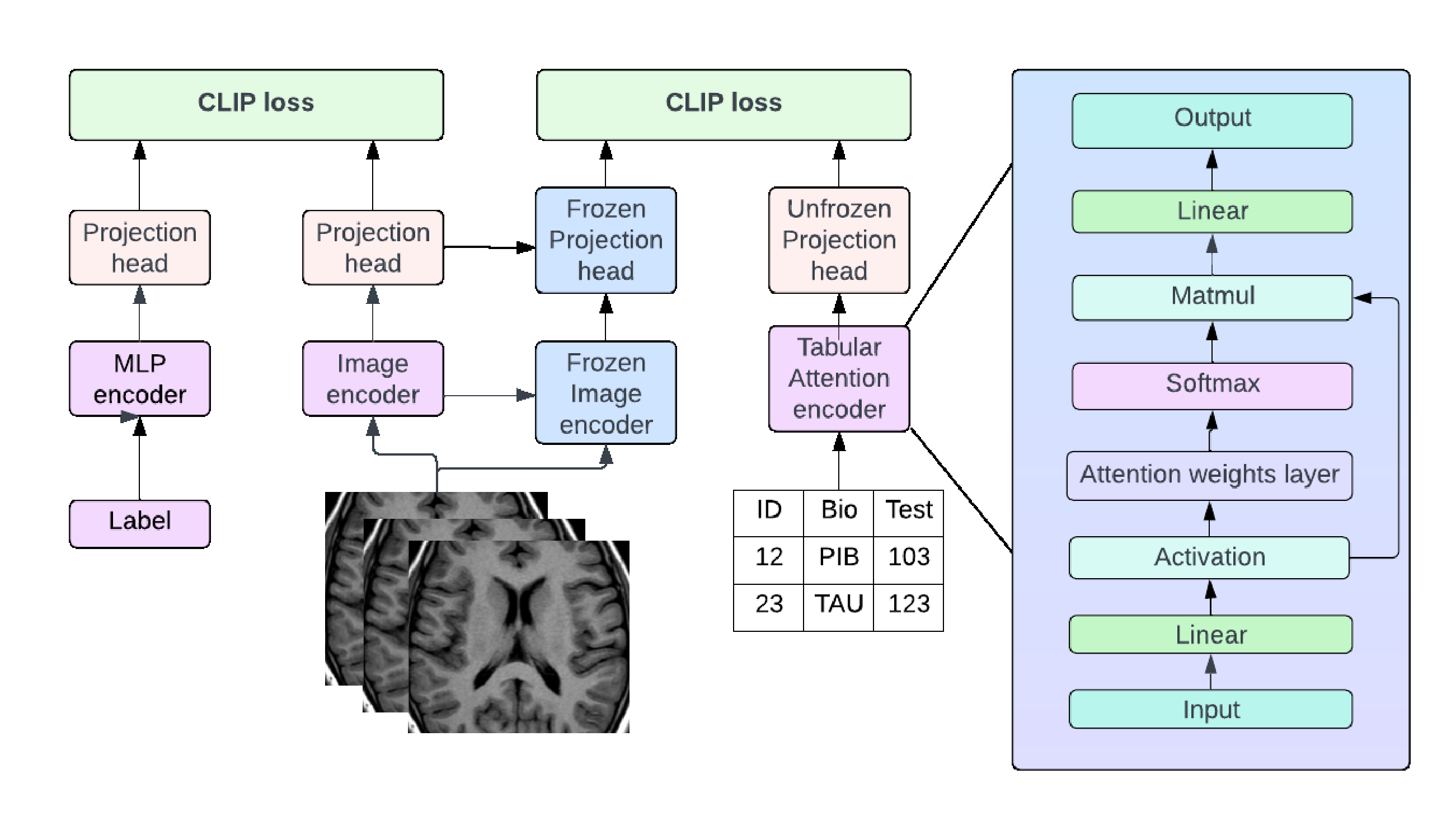}
      \subcaption{Description of the multimodal contrastive learning methodology. We pretrain the image model on the ground-truth label features, and then finetune the tabular data encoders, freezing the image enoder model. We also show the tabular attention module.}
      \label{fig:multimodal_training}
  \end{minipage}
  \hfill
  \begin{minipage}[t]{.45\textwidth}
      \centering
      \includegraphics[width=\linewidth]{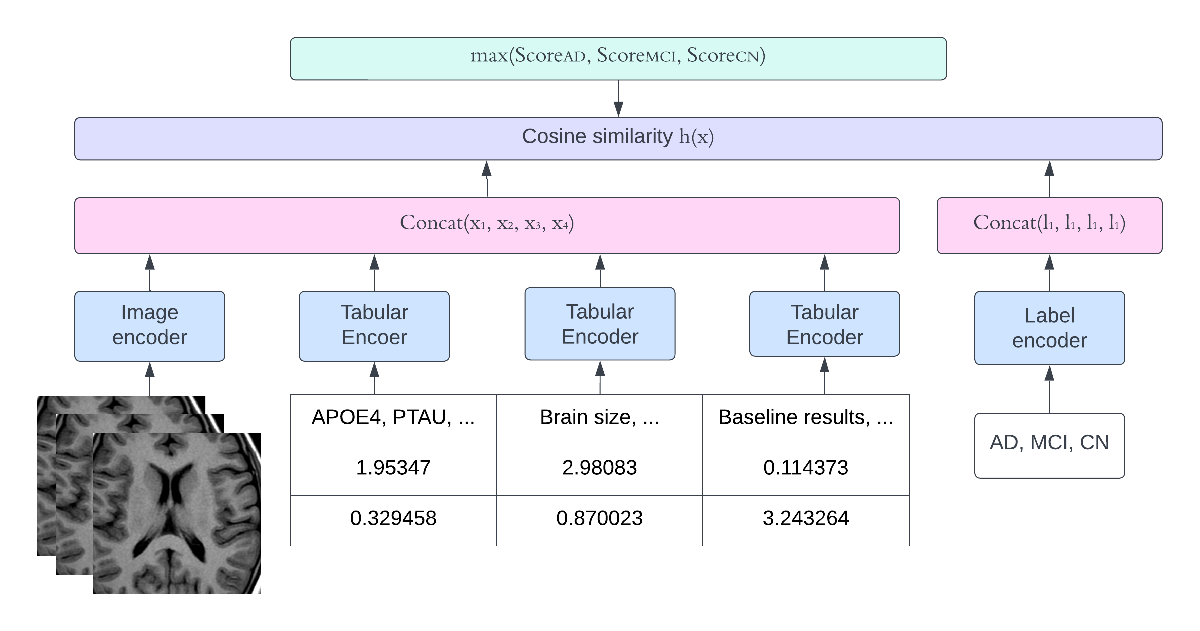}
      \subcaption{Description of the multimodal inference methodology. After encoding each of the features, we concatenate them and compute a cosine similarity. The label with the maximum similarity is the output prediction.}
      \label{fig:multimodal_inference}
  \end{minipage}  
  \label{fig:multimodal_learning}
  \caption{Multimodal Contrastive Learning Framework.}
\end{figure*}

\mycomment{
\begin{figure*}[t!]
  \centering
  \includegraphics[width=0.8\textwidth]{Figs/multimodal_training_v2.eps}
  \caption{Description of the multimodal contrastive learning methodology. We pretrain the image model on the ground-truth label features, and then finetune the tabular data encoders, freezing the image enoder model. We also show the tabular attention module.}
  \label{fig:multimodal_training}
\end{figure*}

\begin{figure*}[t!]
  \centering
  \includegraphics[width=0.6\textwidth]{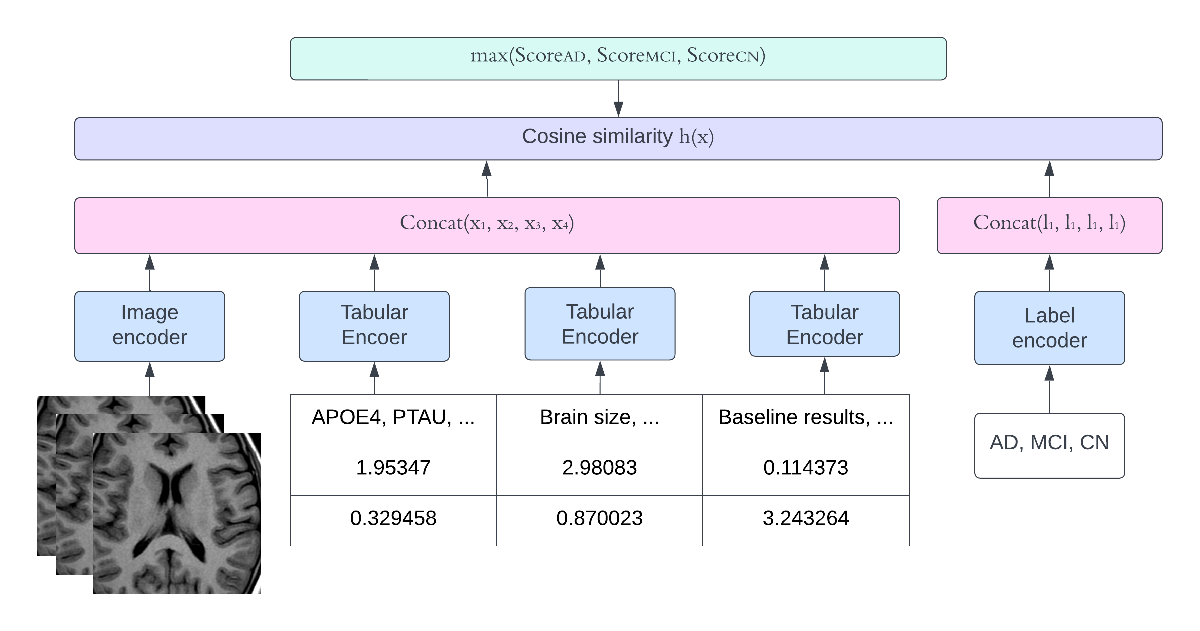}
  \caption{Description of the multimodal inference methodology. After encoding each of the features, we concatenate them and compute a cosine similarity. The label with the maximum similarity is the output prediction.}
  \label{fig:multimodal_inference}
\end{figure*}
}

Alzheimer's disease (AD) is a prevalent neurodegenerative condition affecting millions of individuals worldwide. It has an asymptomatic stage, occurring about 20 years before noticeable symptoms, during which neuronal damage occurs. As the disease progresses, individuals enter the early symptomatic stage characterized by cognitive decline referred to as "mild cognitive impairment due to AD" (MCI). The diagnosis of MCI plays a crucial role in predicting the future development of AD, as approximately 15\% of MCI patients convert to AD annually~\cite{scheltens2016alzheimer, hampel2011future, albert2011diagnosis}.

While there is currently no cure for AD, early diagnosis of the disease holds significant importance. Timely identification allows for the initiation of therapies that can slow down disease progression and improve management strategies~\cite{liu2013apolipoprotein}. In the realm of machine learning and deep learning, previous research has explored the application of deep convolutional networks and structural MR images for the detection of AD and MCI, yielding promising results with accuracy scores of 74.5\%~\cite{song2021effective}.

However, diagnosing AD and MCI solely based on MR images poses challenges, as certain features present in these images, such as cognitive decline, may also be observed in individuals experiencing healthy aging. Furthermore, The recent revised clinical criteria for detecting Alzhiemer's disease \cite{niaaa2023} has cited the importance of using multiple modalities for diagnosis. This study has particularly emphasized the importance of biomarker data, clinical tests, and medical history as primary indications of the disease. However, many of these data types contain disjoint information relating to Alzheimer's. Therefore, we choose to use image data to incorporate the information from these data types into a single framework.

Therefore, it becomes crucial to incorporate additional indicators of MCI and AD into the diagnostic process. These indicators can include diagnostic biomarkers, volumetric data, empirical cognitive assessments, and medical history, among others. Previous works such as~\cite{polsterl2021combining} have shown that multimodal data is crucial to improving the performance of AD and MCI diagnosis. By integrating multiple modalities of data, researchers and clinicians can achieve a more comprehensive and accurate understanding of MCI and AD, improving diagnostic accuracy and treatment planning.

The utilization of multimodal data, beyond MR images alone, enables a holistic approach towards diagnosing and understanding MCI and AD. By considering a combination of imaging data, cognitive assessments, biomarkers, and medical history, healthcare professionals can gain deeper insights into the disease, its progression, and potential treatment options. This multimodal approach addresses the limitations of relying solely on MR images, ensuring a more precise and comprehensive diagnosis of MCI and AD.

Beyond diagnostics, multimodal data is also crucial to advancing our understanding of diseases and the various factors affecting them. Many datasets are multimodal in their nature including~\cite{jack2008alzheimer, azam2022review, johnson2019mimic}. The ADNI~\cite{jack2008alzheimer} includes thousands of tabular data fields and medical examinations, in addition to imaging and genetic information. The inclusion of multiple modalities in the ADNI dataset allows researchers and clinicians to explore various aspects of Alzheimer's disease and gain insights into its underlying properties. For example, the tabular data fields in the ADNI capture a wide range of clinical and demographic information about the participants. This data can include details such as age, sex, education level, medical history, cognitive assessments, and biomarker measurements. By analyzing this information, researchers can identify correlations, risk factors, and disease progression patterns associated with Alzheimer's disease. By integrating and analyzing multimodal data from the ADNI dataset, researchers can unravel complex relationships between clinical, imaging, and genetic factors associated with Alzheimer's disease. This comprehensive approach enhances our understanding of the disease's etiology, progression, and potential therapeutic targets. Furthermore, the ADNI dataset serves as a valuable resource for developing and validating predictive models, biomarkers, and treatment strategies for Alzheimer's and related neurological disorders.

\subsection{Contributions}
To address these issues, we present a generalizable contrastive learning framework, featured with multimodal training as illustrated in Figure~\ref{fig:multimodal_training}, and both multimodal \& unimodal inference capabilities as illustrated in Figure~\ref{fig:multimodal_inference}. To the best of our knowledge, ours is the first generalizable multimodal contrastive learning framework to facilitate the detection of Alzheimer's disease. 

The framework tackles the challenges posed by the diverse range of tabular features, including biomarkers, medical history, and volumetric data, which render classical methods such as CLIP~\cite{radford2021learning} and SimCLR~\cite{chen2020simple} less directly applicable. We adopt a strategy that considers the MR image as the prototypical representation of the patient. Although the image may not capture all the intricacies of an individual's condition, it serves as the shared foundation for all other data types. We train each cluster of tabular features around the image, thereby creating separate contrastive learning tasks that leverage the capabilities of CLIP.

Additionally, we propose a novel attention module for the tabular encoder. This module assigns a score to each column, reflecting the relative importance of the features in the output embedding. By emphasizing the most salient features, the attention mechanism enhances both the overall performance of the model and the intrinsic understanding of the significance of these attributes in the contrastive learning process. Importantly, this approach avoids the need for explicit attribution methods commonly employed in explainable AI~\cite{sundararajan2017axiomatic}.

Our framework not only improves model performance but also illuminates latent relationships that emerge among the various tabular features. By centering the contrastive learning process around the image, these relationships are brought to the forefront, shedding light on the intricate connections that exist between each element.

Figure~\ref{fig:multimodal_training} provides a visual representation of our proposed multimodal training framework, demonstrating how the contrastive learning tasks are organized around the MR image, serving as a central node that unifies the diverse data types and facilitates the exploration of latent relationships.

Overall, our framework offers a promising solution to the challenges posed by multimodal data in contrastive learning, enabling more effective analysis and understanding of complex medical datasets.

\section{Related Work}

\subsection{Contrastive Learning}

Contrastive learning has emerged as a performant successor to pretext tasks. Contrastive learning trains encoders by generating augmented views of a sample and maximizing their projected embedding similarity while minimizing the similarity between the projected embeddings of other samples. It has been popularized by implementations such as SimCLR~\cite{chen2020simple},~\cite{he2020momentum}, BYOL~\cite{grill2020bootstrap}, CLIP~\cite{radford2021learning} and others. We use the contrastive framework of CLIP as the basis for our work.

\subsection{Deep Learning with Tabular Data}

Deep learning has shown competitive performance in analyzing tabular data, but still lags behind simpler algorithms in certain applications~\cite{shwartz2022tabular}. Challenges such as high dimensionality and the need for large labeled datasets affect the effectiveness of deep learning models. Researchers have explored specialized architectures and transfer learning techniques to address these challenges. Ongoing research aims to unlock the full potential of deep learning for tabular data analysis~\cite{borisov2022deep}.

\subsection{Tabular Data in Medical Imaging}

Several works have proposed incorporating tabular data in medical imaging~\cite{zhu2021converting, polsterl2021combining}. Notably,~\cite{hager2023best} introduces a contrastive learning framework with self-supervised pretraining on biobanks. In contrast to these approaches, our method addresses the medical domain specifically and explores the use of the label as a feature~\cite{yang2022unified}. Our approach is tailored for the Alzheimer's detection task, leveraging the image as a prototypical representation for all data for multimodal prediction. Additionally, we incorporate a novel tabular attention to enhance model performance and interpretability. 

While previous works primarily focus on unimodal prediction, our approach enables both unimodal and  multimodal prediction. Furthermore, we tackle the challenge of selecting the best features without relying on explainable AI methods~\cite{sundararajan2017axiomatic}. We extend the few-shot learning setting by removing the need for self-supervised pretraining and utilize a ResNet model for enhanced performance. Our approach also considers categorical feature encoding and includes MRI as an essential data modality. Through these advancements, our method improves upon existing approaches, providing a comprehensive framework for multimodal Alzheimer's detection in the medical domain.

\subsection{Multimodal Data in Alzheimers's Prediction}

In the realm of multimodal data analysis, several studies have focused on the ADNI dataset, including~\cite{golovanevsky2022multimodal}, which introduces a cross-modal attention-based solution. However, these existing approaches often lack a generalizable structure for multimodal learning and are limited by fixed data modalities. Consequently, significant modifications are required to incorporate different types of data into their model architectures. Moreover, the aforementioned work primarily focuses on the binary classification task of distinguishing between AD and MCI, overlooking the inclusion of CN (normal) cases. 

In contrast, our proposed approach extends beyond binary classification and can effectively detect AD vs CN on a spectrum, providing a more comprehensive diagnostic capability. In terms of generalizability, our method offers a more flexible framework for multimodal learning. By designing a modular architecture, our model can easily incorporate diverse data modalities without requiring substantial modifications. This adaptability allows for seamless integration of additional data types and facilitates future extensions to encompass emerging modalities. Furthermore, our approach leverages the novel attention module for the tabular encoder, enabling the model to assign importance scores to each column. This feature enhances the interpretability of the model by highlighting the most relevant features for classification, which is especially crucial in medical diagnosis tasks where feature importance and transparency are highly valued.

\section{Methodology}

\subsection{Problem Formulation}

We define the multimodal Alzheimer's disease prediction as a contrastive learning problem between multiple modalities. The use of contrastive learning is motivated by the fact that the image and tabular data is correlated. We use the contrastive learning framework to learn a joint embedding space for the modalities. The joint embedding space is then used to calculate similarities between embeddings of a particular label and the embeddings of our input modalities. The similarity scores are then used to predict the label of the input modalities. We leverage the image as a prototypical modality to learn the joint embedding space. The image modality is chosen as the prototypical modality because it is the most expressive modality. The image modality is used to learn the joint embedding space by using a contrastive learning framework. We align
each modality's embedding to image embeddings, such as biomarker to image using a tabular encoder. We show that the resulting embedding space has a powerful emergent zero-shot behavior that automatically associates pairs of modalities without seeing any training data for that specific pair.

\subsection{Contrastive Learning Method}

We use pairs of modalities $(I, M)$, where $I$ represents the MR image and $M$ denotes a tabular modality, to learn a single joint embedding space. Consider a specific pair of modalities $(I, M)$ with aligned features. Given an MR image $I_i$ and its corresponding tabular value $M_i$, we encode them into normalized embeddings: $q_i=f(I_i)$ and $k_i=g(M_i)$. Here, the function $f$ represents a convolutional encoder for processing the MR image, while $g$ denotes a tabular encoder for handling the tabular modality. 

To optimize the embeddings and the encoders, we employ the CLIP (Contrastive Language-Image Pretraining) loss~\cite{radford2021learning}. The CLIP loss functions to align the embeddings of similar pairs and push apart the embeddings of dissimilar pairs. By maximizing the similarity between corresponding MR image and tabular embeddings, we encourage the joint embedding space to capture the shared information between the modalities effectively. Conversely, we minimize the similarity between embeddings from different MR image and tabular pairs, ensuring that unrelated pairs are separated in the joint embedding space.

The mathematical formulation for the CLIP loss can be expressed as:

\[
\mathcal{L}_{\text{CLIP}} = -\log\frac{\exp(q_i \cdot k_i / \tau)}{\sum_{j=1}^{N}\exp(q_i \cdot k_j / \tau)}
\]

where $N$ denotes the total number of pairs, $q_i$ and $k_i$ are the normalized embeddings for the MR image and tabular modality respectively, $i$ and $j$ represent indices that iterate over pairs of embeddings, and $\tau$ is a temperature parameter that controls the sharpness of the similarity distribution. By minimizing this loss, we encourage the embeddings of similar pairs to have higher cosine similarity scores while pushing the dissimilar pairs towards lower similarity scores. By leveraging the joint embeddings and optimizing them using the CLIP loss, our model learns to effectively align and fuse information from both the MR image and tabular modality.

Furthermore, by contrasting modalities with images, we observe emergent alignment of unseen pairs of modalities. Given $(I, M_1)$ and $(I, M_2)$ where $M_1$ and $M_2$ are distinct modalities, there exists emergent behavior when comparing the pair $(M_1, M_2)$ even though the encoders were only trained on $(I, M_1)$ and $(I, M_2)$. This behavior allows us to combine predictions for multimodal training and inference. 

\subsection{Tabular Attention}

We propose a modified architecture for incorporating tabular attention in our multimodal Alzheimer's disease detection model. Traditional attention mechanisms are designed to compute the relevance or importance of different elements in a sequence relative to each other. However, when dealing with a single row of tabular data, there are no other elements to compare or attend to within the sequence.

To address this limitation, we adopt an alternative approach that allows us to highlight specific elements within a row and assign them different weights without using a traditional attention mechanism. Instead, we utilize a learnable weight matrix or a simple gating mechanism to control the emphasis on specific elements in the row.

In this modified architecture, we introduce a step to initialize the weight matrix or gating mechanism specifically for tabular attention. This mechanism allows us to assign varying weights to different elements within the row, emphasizing specific elements of interest. After applying the tabular attention mechanism to the row, we continue with the remaining layers of the model to process the attended row and extract relevant features. Finally, the predicted class is obtained based on the output of the model.

By incorporating this modified architecture with tabular attention, we aim to enhance the model's ability to capture important features within the tabular data, leading to improved performance in Alzheimer's disease detection. Furthermore, the attention weights $W$ can be used to rank the importance of each column in the table, which allows for more interpretability.

\begin{algorithm}[H]
\caption{Modified Architecture with Tabular Attention}
\begin{algorithmic}[1]
\Require Tabular data row $X$
\Ensure Predicted class $\hat{y}$

\State Initialize the weight matrix or gating mechanism for tabular attention
\State Compute the attention weights $W$ for each element in $X$ using the weight matrix or gating mechanism

\State Apply the attention weights to the tabular row: $X_{\text{attended}} = X \odot W$, where $\odot$ denotes element-wise multiplication

\State Pass the attended row $X_{\text{attended}}$ through the remaining layers of the model
\State Compute the final output $O$ based on the processed attended row

\State Obtain the predicted class $\hat{y}$ based on $O$ (e.g., through softmax or thresholding)

\State \textbf{return} $\hat{y}$
\end{algorithmic}
\end{algorithm}

\subsection{Multimodal Inference and Evaluation}

We evaluate the performance of our model by assessing its ability to accurately predict Alzheimer's disease (AD) vs cognitively normal (CN) cases. However, the versatility of our framework extends beyond this specific task, allowing for its application to various other scenarios.

As mentioned previously, our contrastive learning framework can accommodate both unimodal and multimodal inputs. For instance, given input data $(I, L, M_1, M_2)$, where $L$ represents a potential label, the similarity function $h(x, y)$ can be defined for $x = [I, M_1, M_2]$ and $y = [L]$. In this case, the features can be concatenated, and the cosine similarity can be computed as follows:

\[
h(x, y) = \frac{{x \cdot y}}{{\|x\| \cdot \|y\|}}
\]

This similarity function allows us to measure the degree of similarity or dissimilarity between the combined multimodal features $x$ and the label feature $y$. The label with the maximum similarity is the output prediction.

Since we defined the labels as a spectrum, we can apply probabalistic ternary search to search for the $y$ value with highest similarity in logarithmic time complexity:

\begin{algorithm}
\caption{Probabilistic Ternary Search for Maximum Similarity}
\begin{algorithmic}[1]
\Require Sorted array $y$ of label values, similarity function $h(x, y)$, target similarity threshold $T$
\Ensure Predicted label $\hat{y}$ with maximum similarity

\State Set low and high indices: $l = 0$, $h = \text{len}(y) - 1$

\While {$h - l > 2$}
    \State Set mid1 and mid2 indices: $m1 = \lfloor l + (h - l) / 3 \rfloor$, $m2 = \lfloor h - (h - l) / 3 \rfloor$
    
    \State Calculate similarities: $s1 = h(x, y[m1])$, $s2 = h(x, y[m2])$
    
    \If {$s_1 > s_2$}
        \State Set new high index: $h = m2$
    \Else
        \State Set new low index: $l = m1$
    \EndIf
\EndWhile

\State Calculate remaining similarities: $s_l = h(x, y[l])$, $s_h = h(x, y[h])$

\If {$s_l > s_h$}
    \State \textbf{return} $y[l]$
\Else
    \State \textbf{return} $y[h]$
\EndIf
\end{algorithmic}
\end{algorithm}

In this algorithm, we perform a probabilistic ternary search on the sorted array of label values $y$ to find the label $\hat{y}$ with the maximum similarity to the combined multimodal features $x$. We initialize the low and high indices as the first and last indices of $y$, respectively.

During each iteration of the while loop, we calculate the similarities $s_1$ and $s_2$ for the mid1 and mid2 indices. We compare the similarities and update the low or high index accordingly to narrow down the search space.

Once the while loop terminates and only three elements remain, we calculate the remaining similarities $s_l$ and $s_h$ for the low and high indices. Finally, we compare $s_l$ and $s_h$ to determine the label with the highest similarity, and return the predicted label $\hat{y}$.

By applying this probabilistic ternary search algorithm, we efficiently identify the progression of the disease that exhibits the maximum similarity to the combined multimodal features $x$, allowing for accurate predictions in our multimodal Alzheimer's disease detection framework.

Moreover, our framework is not limited to only multimodal inputs. If the length of $x$ is 1, indicating a unimodal input, we can still leverage the contrastive learning framework effectively. In this scenario, the contrastive loss encourages the model to distinguish between different instances of the same modality, facilitating the learning of discriminative representations within the unimodal data. By incorporating both unimodal and multimodal inputs, our framework provides a flexible and adaptable solution that can be applied to a wide range of tasks and data types, promoting comprehensive learning and improved performance in various applications. 

A diagram for this process is shown in Figure~\ref{fig:multimodal_inference}.

\subsection{Implementation Details}

\subsubsection{Dataset}

We utilize the Alzheimer's Disease Neuroimaging Initiative (ADNI) dataset, which is a comprehensive collection of multimodal data from subjects with normal cognition, mild cognitive impairment (MCI), and Alzheimer's disease (AD)~\cite{jack2008alzheimer}. To extract the tabular data from ADNI, we employ the ADNIMerge tool, which consolidates various data sources into a unified format for analysis. The tabular data consists of diverse columns, which we categorize into several categories to capture different aspects of the patients' information. These categories include biomarkers such as APOE4 and pTau, which provide genetic and protein level information~\cite{montagne2020apoe4}, empirical cognitive assessments including MMSE and RAVLT~\cite{arevalo2015mini}, volumetric data like hippocampus size and brain size, and medical history details such as the baseline diagnosis.

To handle categorical features, we apply one-hot encoding, transforming each categorical variable into a binary representation. This encoding scheme allows us to incorporate categorical information into our models effectively.

For the label encoding, we adopt a spectrum-based approach. We assign numerical values to the labels as follows: CN (cognitively normal) is encoded as 0, MCI (mild cognitive impairment) as 0.5, and AD (Alzheimer's disease) as 1. This spectrum-based encoding enables us to capture the progression of cognitive impairment, providing a more nuanced representation of the patients' conditions. For instance, a patient in the late stage of MCI may receive a score of 0.75, indicating their intermediate position between MCI and AD.

To ensure fair comparisons and facilitate convergence during training, we apply z-score normalization to the numerical features. This normalization technique rescales the values to have a mean of zero and a standard deviation of one. By normalizing the numerical features, we mitigate the influence of the features with larger magnitudes and ensure that each feature contributes equally during model training.

For the experiments, we partition the data into training, validation, and test sets. We use a 70-15-15 split. The training set is used to train the model, the validation set is used to tune the hyperparameters, and the test set is used to evaluate the model's performance. We use the same splits for all experiments to ensure consistency.

\subsubsection{Training Procedures}

In this section, we describe the training procedure employed for our multimodal Alzheimer's disease detection model. We encode all parts of the data into features, including labels, to ensure comprehensive information representation during training. To initialize the model, we adopt a contrastive learning formulation and pretrain both the image and label encoders. This pretraining step enables the model to learn rich representations from the MR image data and label information, enhancing the effectiveness of subsequent training steps.

During the main training phase, we employ a ResNet~\cite{he2016deep} as our encoder with a projection size of 128. The ResNet architecture allows the model to capture spatial features and patterns in the MR images effectively. For the label encoder, we utilize a Multilayer Perceptron (MLP)~\cite{amari1967theory} architecture to encode the label information, enabling the model to capture the semantic representation of the disease states. To handle the tabular data, we incorporate tabular attention as the tabular encoder. This attention mechanism assigns importance scores to each column, allowing the model to focus on the most relevant features during training and prediction. The attention module enhances the interpretability and performance of the model.

For all experiments, we train the model for 64 epochs using the training data. We choose this epoch duration to ensure sufficient training iterations while avoiding overfitting. During training, we monitor the validation loss, and we select the model checkpoint with the lowest validation loss for subsequent evaluation and testing. In our training process, we adopt distinct batch sizes depending on the dimensionality of the images. Specifically, a batch size of 4 is employed for 3D images, while a batch size of 32 is utilized for 2D images. To optimize the model's performance, we integrate the Adam optimizer with a learning rate set at 0.0001, accompanied by a weight decay of 0.01 to control overfitting. Additionally, we incorporate a learning rate scheduler, which employs a decay rate of 0.1 and a patience of 10 epochs, contributing to the gradual adjustment of the learning rate during training. With a focus on efficiency, our training spans 64 epochs, incorporating early stopping to prevent unnecessary iterations. Finally, for the CLIP loss, a temperature of 0.1 is employed, ensuring an appropriate balance between optimization and exploration.

\subsubsection{Evaluation}

To evaluate the performance and effectiveness of our models, we adhere to the inference structure outlined above, focusing on two important classification tasks: distinguishing between individuals with normal cognition (CN) and those with Alzheimer's disease (AD), as well as discerning between CN, mild cognitive impairment (MCI), and AD. The evaluation process allows us to gain insights into the accuracy and robustness of our models in classifying these different cognitive states. In order to assess the performance of our models, we employ a label-balanced split of 882 MR image slices. This ensures that each cognitive state is represented proportionally within the evaluation dataset, providing a fair and representative sample for assessment purposes. By including an ample number of MR image slices, we ensure that our evaluation captures a comprehensive view of the model's performance across various cases and scenarios. We run the model 5 times across different testing splits and report the mean and standard deviation of the accuracy.

\section{Results}

\subsection{Ablation Study}

Table 1 presents the accuracy of the model for different data modalities in predicting Alzheimer's disease (AD) versus cognitively normal (CN) cases and overall accuracy for all labels. MR images achieve the highest accuracy (88.5\%) in distinguishing AD from CN cases, capturing structural brain changes associated with the disease. Cognitive tests also demonstrate high accuracy (91.4\%), providing valuable information about cognitive decline. Medical history achieves a high accuracy (92.5\%) by incorporating risk factors and comorbidities. However, biomarkers and volumetric data achieve relatively lower accuracies (68.7\% and 82.1\%) possibly due to noise and limited complexity capture. The multimodal input method achieves the highest accuracy (95.5\% for AD vs CN and 83.8\% for All Labels Accuracy) by combining all data modalities. This result demonstrates the effectiveness of the framework in capturing complex relationships between different data types, leading to improved Alzheimer's disease detection. Moreover, the multimodal approach achieves the highest accuracy for all labels, indicating its effectiveness in capturing diverse patterns. These results underscore the importance of multimodal data in enhancing model performance and improving Alzheimer's disease detection accuracy.

It is interesting to note that medical history alone outperforms image data. Medical history contains a wide variety of variables, including baseline diagnosis results, demographic information, and cognitive test scores. These variables are highly correlated with the progression of Alzheimer's disease, and therefore, the medical history modality is able to provide a comprehensive overview of the patient's condition. Statistically, many of these indications lead to Alzhiemer's disease. On the other hand, this data does not fully capture the progression of the disease, as it is only a snapshot of the patient's condition at a particular point in time. Therefore, the image modality is able to provide additional information that is not captured by the medical history modality. By combining the information from both modalities, our model is able to leverage the strengths of each modality to achieve the best performance.

\subsection{Comparison with Previous Methods}

Table 2 compares different model types. The proposed models incorporating tabular attention with contrastive learning outperform previous methods by 9.3\% over the state of the art, benefiting from attention's ability to focus on relevant tabular features. Additionally, the proposed models outperform the previous tabular method by 21.6\% in accuracy, demonstrating the effectiveness of contrastive learning in facilitating Alzheimer's disease detection.

\begin{table}[h]
\label{tab:accuracy}
\begin{tabular}{|p{3cm}|p{2cm}|p{2cm}|}
\hline
\textbf{Data Modality} & \textbf{AD vs CN Accuracy} & \textbf{All Labels Accuracy} \\
\hline
MR Images & 0.885 $\pm$ 0.015 & 0.761 $\pm$ 0.014 \\
\hline
Biomarkers & 0.687 $\pm$ 0.159 & 0.428 $\pm$ 0.057 \\
\hline
Cognitive Tests & 0.914 $\pm$ 0.061 & 0.758 $\pm$ 0.024 \\
\hline
Volumetric Data & 0.821 $\pm$ 0.051 & 0.516 $\pm$ 0.036 \\
\hline
Medical History & 0.925 $\pm$ 0.865 & 0.789 $\pm$ 0.041 \\
\hline
\textbf{Multimodal (All Data)} & \textbf{0.955 $\pm$ 0.017} & \textbf{0.838 $\pm$ 0.023} \\
\hline
\end{tabular}
\caption{Mean accuracy and standard deviation of different testing modalities.}
\end{table}

\begin{table}[h]
\label{tab:accuracy}
\begin{tabular}{|p{5.4cm}|p{2cm}|}
\hline
\textbf{Model Type} & \textbf{Accuracy} \\
\hline
DAFT Tabular Fusion~\cite{polsterl2021combining} & 0.622 \\
\hline
3D CNN Image Fusion~\cite{song2021effective} & 0.745 \\
\hline
\textbf{Tabular MLP Encoder + Contrastive} & \textbf{0.799} \\
\hline
\textbf{Tabular Attention + Contrastive} & \textbf{0.838} \\
\hline
\end{tabular}
\caption{Accuracy of different model types, comparing with previous methods.}
\end{table}

\begin{figure}[ht]
  \centering
  \includegraphics[width=\linewidth]{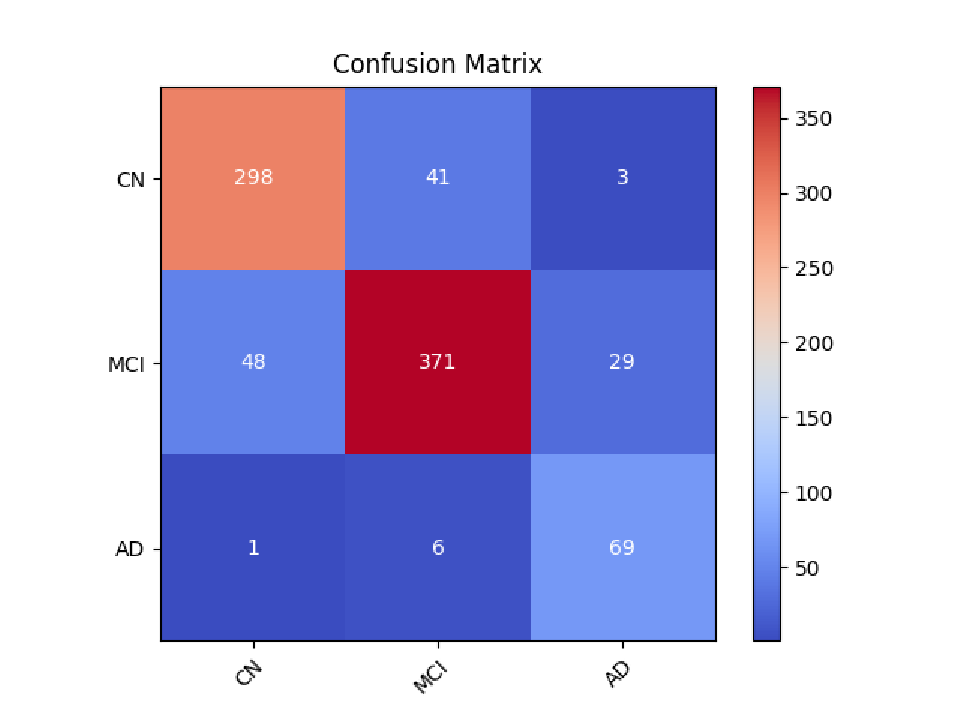}
  \caption{Confusion Matrix}
  \label{fig:confusion_matrix}
\end{figure}

\begin{figure*}[p] 
  \begin{minipage}[t]{0.5\linewidth}
      \centering
      \includegraphics[width=\linewidth]{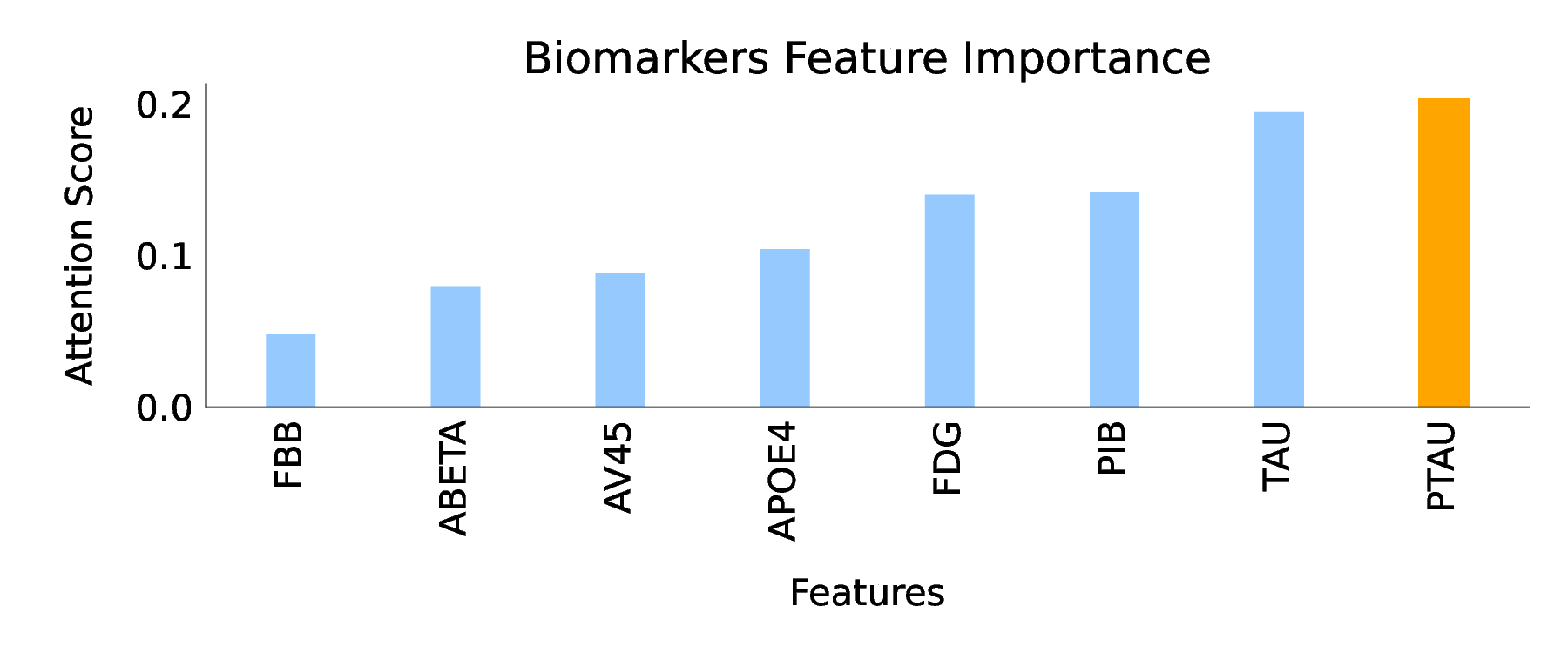}
      \caption{Attention Scores for Biomarkers}
      \label{fig:biomarkers}
  \end{minipage}
  \hfill
  \begin{minipage}[t]{0.5\linewidth}
      \centering
      \includegraphics[width=\linewidth]{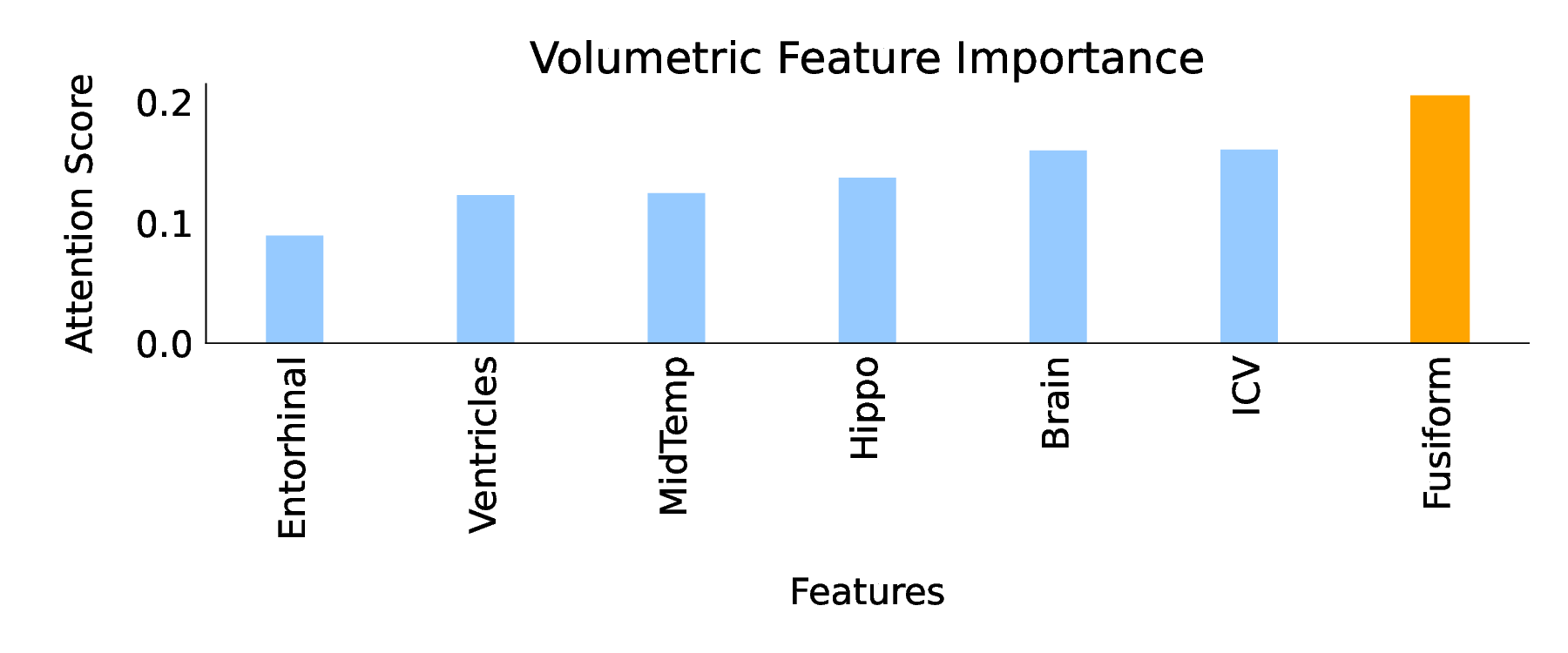}
      \caption{Attention Maps for Volumetric Data of the Brain}
      \label{fig:volumetric}
  \end{minipage}
  \begin{minipage}[t]{\linewidth}
    \centering
    \includegraphics[width=\linewidth]{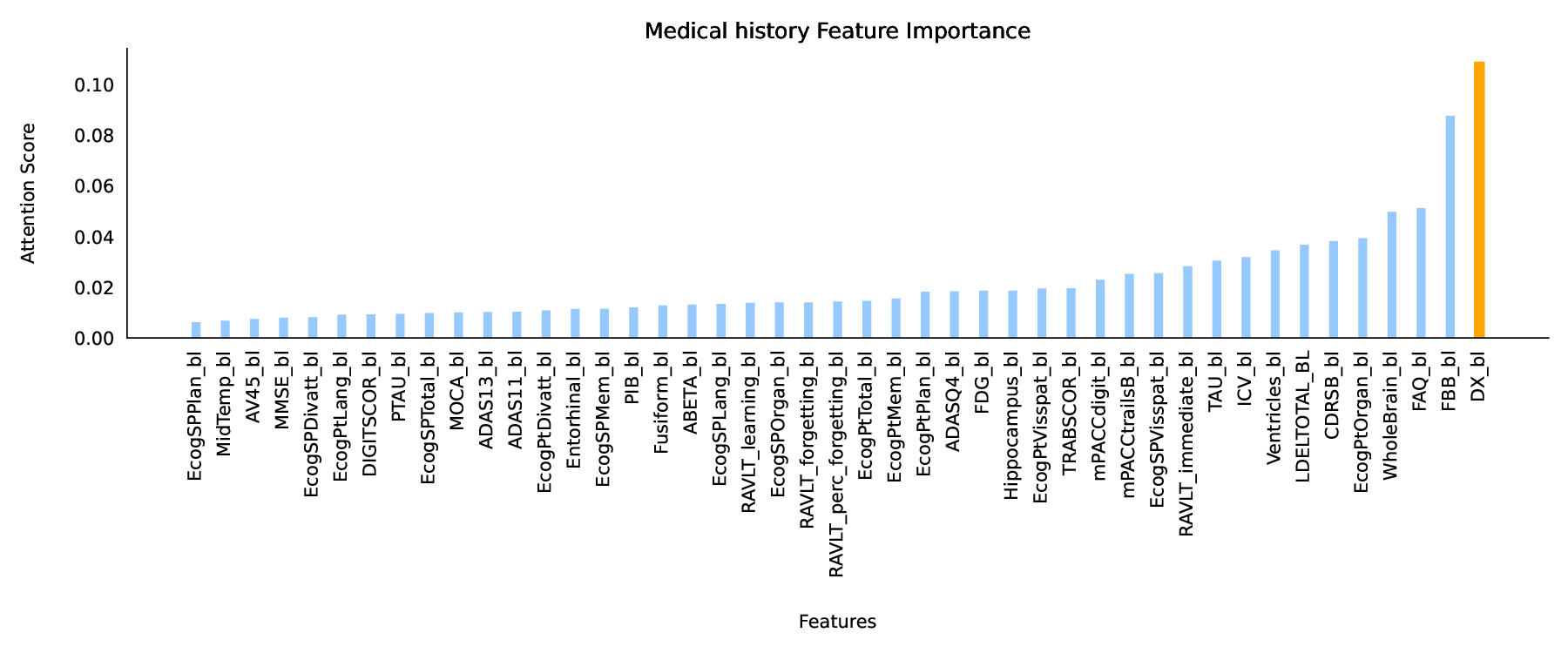}
    \caption{Attention Scores for Medical History}
    \label{fig:medical}
  \end{minipage}
  \hfill
  \begin{minipage}[t]{\linewidth}
      \centering
      \includegraphics[width=\linewidth]{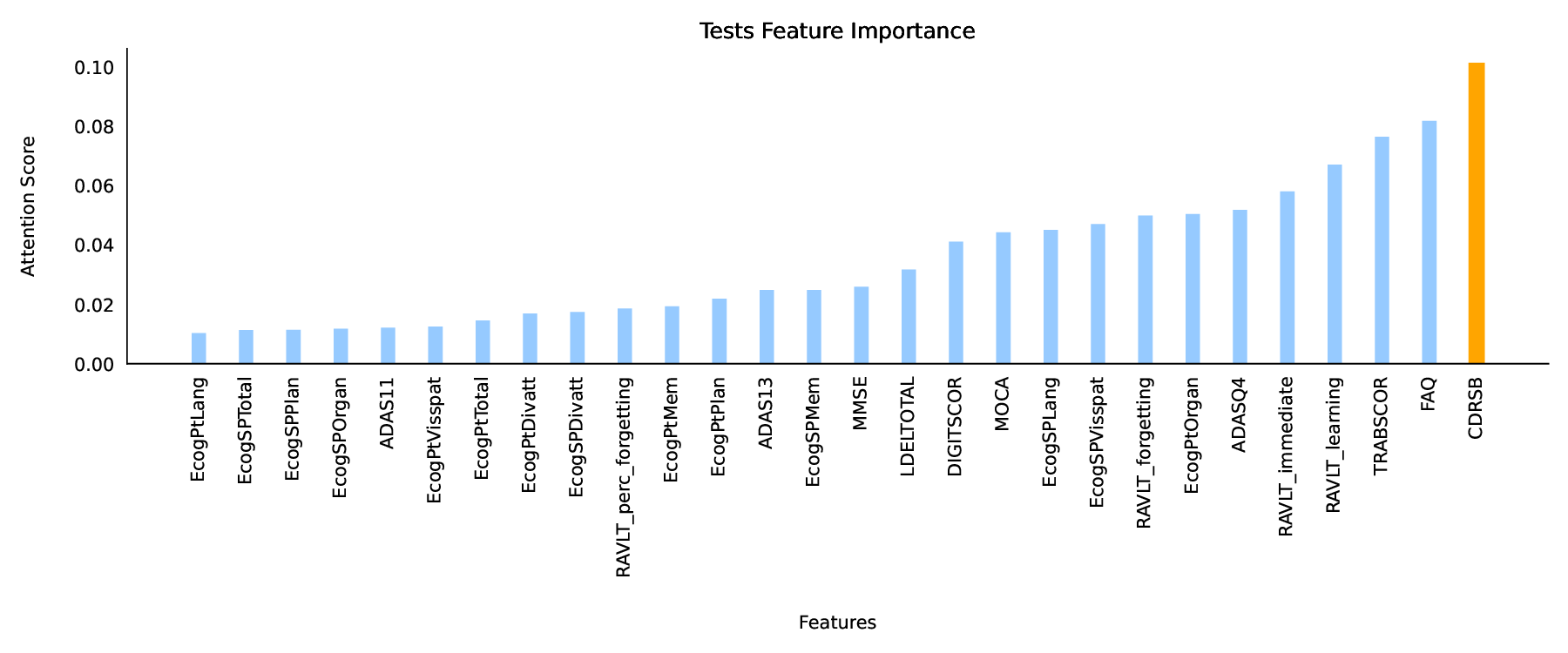}
      \caption{Attention Maps for Cognitive Tests}
      \label{fig:tests}
  \end{minipage}
\end{figure*}

\subsection{Attention Maps}

Figure~\ref{fig:biomarkers} shows the attention scores assigned to each biomarker. The PIB-PET-derived beta-amyloid signature exhibits the highest scores, suggesting its significant influence in predicting metabolic, gray matter, and cognitive changes in non-demented subjects. PIB-PET scans visualize and quantify beta-amyloid plaques in the brain. The high attention score for the PIB-PET-derived beta-amyloid signature indicates its relevance in predicting metabolic, gray matter, and cognitive changes, and its role in Alzheimer's disease-related neurodegenerative processes~\cite{ewers2012csf}. These biomarkers serve as informative markers for detecting and monitoring early-stage cognitive decline and Alzheimer's disease progression in non-demented individuals. Understanding their significance can help identify individuals at higher risk of dementia and guide the development of targeted interventions and therapeutic strategies.

Figure~\ref{fig:volumetric} displays the attention maps for brain volumetric data. These maps highlight regions of interest and focus on various volumetric features. The most prominent features are the whole brain size and Total Intracranial Volume (TIV), followed by the entorhinal volume. Research has extensively studied the entorhinal cortex volume concerning episodic memory and brain activation in normal aging and amnesic mild cognitive impairment (MCI). The entorhinal cortex plays a crucial role in memory formation and retrieval processes. The attention map's emphasis on the entorhinal volume indicates its significance in capturing brain activation related to episodic memory, consistent with prior research linking it to memory-related brain functions. Additionally, the attention map underscores the importance of whole brain size and Total Intracranial Volume~\cite{devanand2007hippocampal}, providing a measure of the overall brain structure and size. The high attention scores assigned to these features suggest their relevance in characterizing brain changes associated with normal aging and cognitive impairment. Changes in brain size, as indicated by these features, can reflect global structural alterations and may indicate neurodegenerative processes.

Figures~\ref{fig:medical},~\ref{fig:tests} demonstrate attention maps for cognitive tests and medical history, providing insights into crucial factors for diagnosing and evaluating Alzheimer's disease (AD). 

Figure~\ref{fig:medical} presents attention scores for various aspects of medical history, with two factors standing out with the highest scores: whole brain size and baseline diagnosis. The brain's overall size, measured through neuroimaging techniques, is of great interest in AD research due to its correlation with brain atrophy, a characteristic feature of the disease~\cite{scahill2002mapping}. Additionally, the baseline diagnosis plays a crucial role in understanding patients' initial cognitive status and serves as a reference point for monitoring disease progression over time.

Figure~\ref{fig:tests} shows attention scores for cognitive tests, with the Clinical Dementia Rating-Sum of Boxes (CDRSB) test receiving the highest score. The CDRSB is widely recognized as a reliable tool for assessing dementia severity and cognitive impairment in AD patients~\cite{morris1993clinical}. Its high attention score emphasizes its significance in the diagnostic process and evaluation of disease progression.

Understanding these attention maps helps identify key brain regions and features relevant to cognitive functioning and decline, facilitating early detection and monitoring of cognitive impairments.

The confusion matrix in Figure~\ref{fig:confusion_matrix} provides a comprehensive analysis of the classification performance of our model on the dataset. The matrix presents the classification results for three classes: CN (Cognitively Normal), MCI (Mild Cognitive Impairment), and AD (Alzheimer's Disease). The rows of the matrix represent the actual labels, while the columns represent the predicted labels. The values within the matrix indicate the number of samples assigned to each category. Our model achieved high accuracy in predicting CN (298 out of 342) and MCI (371 out of 448) cases. However, there were misclassifications, particularly in distinguishing between MCI and AD, with 48 MCI samples being classified as AD. These results provide insights into the strengths and weaknesses of the model, highlighting areas where further improvements can be made to enhance the accuracy and minimize misclassifications.

\section{Conclusion and Future Works}

This paper presents a comprehensive and generalizable framework for multimodal Alzheimer's disease detection, leveraging contrastive learning and attention mechanisms. The results demonstrate the effectiveness of our approach in accurately predicting AD, MCI, and CN labels using different data modalities. By integrating multiple modalities, including MR images, biomarkers, cognitive tests, volumetric data, and medical history, our multimodal framework achieves superior performance compared to individual modalities alone. The multimodal approach and attention mechanism provide a more comprehensive representation of Alzheimer's disease, capturing diverse patterns and improving the accuracy of predictions across different labels.

Furthermore, the incorporation of tabular attention in our model showcased the significance of specific features within the tabular data. The attention mechanism highlights the importance of biomarkers, such as the CSF biomarker and PIB-PET-derived beta-amyloid signature, in predicting metabolic, gray matter, and cognitive changes. Additionally, the attention maps for volumetric data emphasize the relevance of features like whole brain size, total intracranial volume, and entorhinal volume in characterizing brain alterations associated with Alzheimer's disease. These findings have important implications for the early detection and monitoring of Alzheimer's disease. 

Future research can further expand the framework to include other neurodegenerative diseases. Furthermore, the interpretability of the attention mechanisms can be explored to provide insights into the underlying mechanisms and relationships within the multimodal data, possibly enabling the model to perform a much wider range of predictive tasks. 

{\small
\bibliographystyle{unsrt}
\bibliography{arxiv}
}

\end{document}